\documentclass{article}
\usepackage{ijcai24}

\usepackage{times}
\usepackage{soul}
\usepackage{url}
\usepackage[hidelinks]{hyperref}
\usepackage[utf8]{inputenc}
\usepackage[small]{caption}
\usepackage{graphicx}
\usepackage{amsmath}
\usepackage{amsthm}
\usepackage{booktabs}
\usepackage{algorithm}
\usepackage{algorithmic}
\usepackage[switch]{lineno}

\usepackage{multirow}

\hypersetup{
    colorlinks=true,       
    urlcolor=magenta,
    citecolor=black,
    linkcolor=red,   
}

\title{Generating Multi-Center Classifier via Conditional Gaussian Distribution}

\author{
Zhemin Zhang$^1$
\and
Xun Gong$^1$
\affiliations
$^1$Southwest Jiaotong University, China
\emails
zheminzhang@my.swjtu.edu.cn
}

\begin{document}

\maketitle

\begin{abstract}
The linear classifier is widely used in various image classification tasks. It works by optimizing the distance between a sample and its corresponding class center. However, in real-world data, one class can contain several local clusters, \emph{e.g.}, birds of different poses. To address this complexity, we propose a novel multi-center classifier. Different from the vanilla linear classifier, our proposal is established on the assumption that the deep features of the training set follow a Gaussian Mixture distribution. Specifically, we create a conditional Gaussian distribution for each class and then sample multiple sub-centers from that distribution to extend the linear classifier. This approach allows the model to capture intra-class local structures more efficiently. In addition, at test time we set the mean of the conditional Gaussian distribution as the class center of the linear classifier and follow the vanilla linear classifier outputs, thus requiring no additional parameters or computational overhead. Extensive experiments on image classification show that the proposed multi-center classifier is a powerful alternative to widely used linear classifiers. Code available at \url{https://github.com/ZheminZhang1/MultiCenter-Classifier}.
\end{abstract}

\section{Introduction}

Recently, deep neural networks have made significant advancements in various computer vision classification tasks, such as image object recognition \cite{Liu-2021-ICCV,Ren-2023-ICCV,Dong-2022-CVPR}, face recognition \cite{Deng-2019-CVPR,Yu-2023-ICCV}, target detection \cite{wang2023goldyolo,Zhu-2021-ICCV}, fine-grained image classification \cite{Kotovenko-2023-CVPR,Ermolov-2022-CVPR}, etc. Trained on large-scale image data, deep neural networks effectively replace manually engineered feature extractors, demonstrating the superior ability in extracting highly discriminative features from massive data. 

\begin{figure}[t]
\centering
\includegraphics[width=0.95\linewidth]{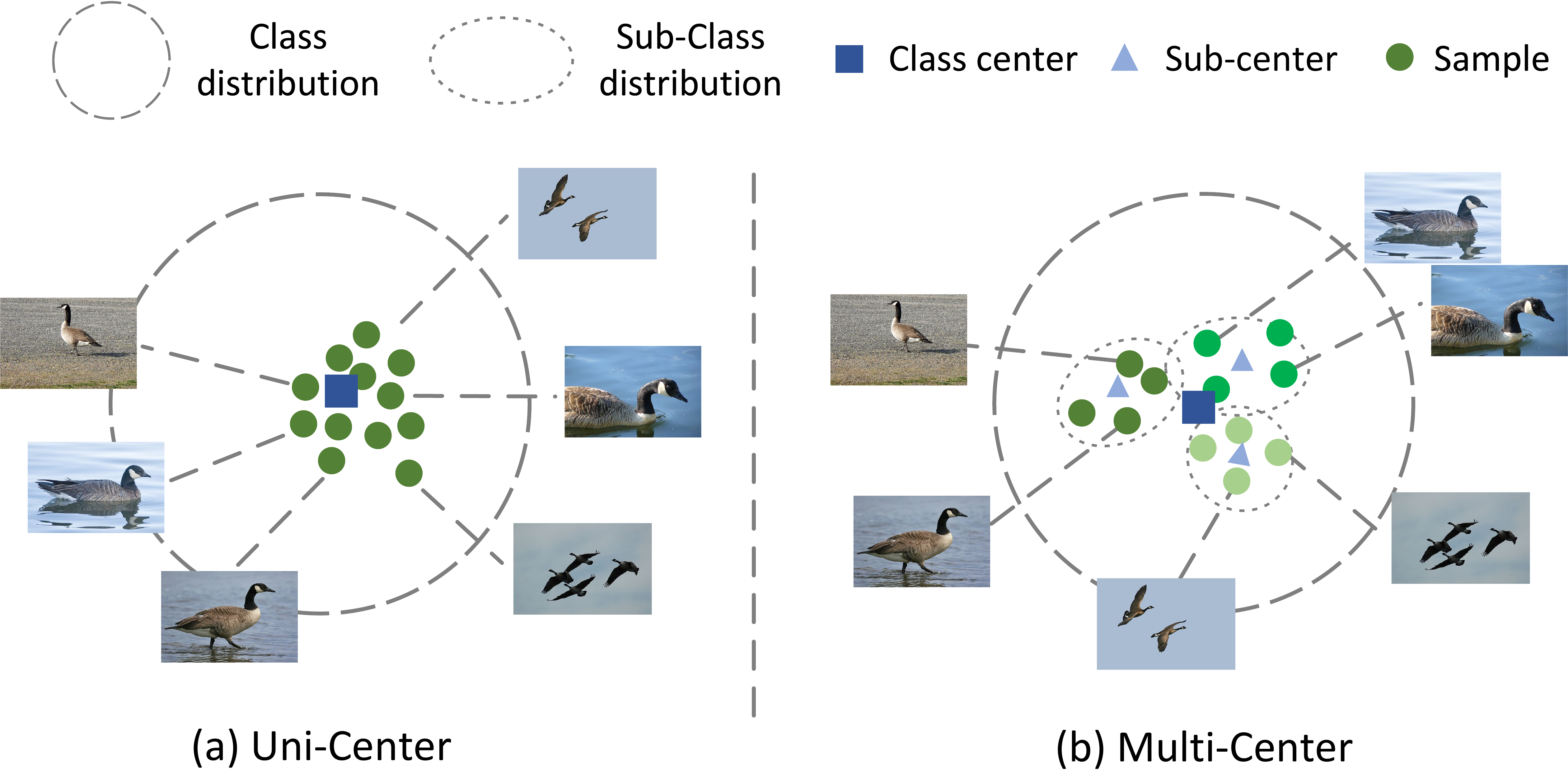} 
\caption{The comparison between uni-center and multi-center approaches when dealing with classes that have different sub-classes. (a) In the uni-center approach, samples belonging to the same class are assigned to a single center, which may not be suitable for real-world data. (b) Conversely, the multi-center approach allows for greater flexibility in modeling intra-class variance by setting multiple sub-centers within a class.}
\label{MultiCenterImageEg-flabel}
\end{figure}

\begin{figure*}[t]
\centering
\includegraphics[width=0.8\linewidth]{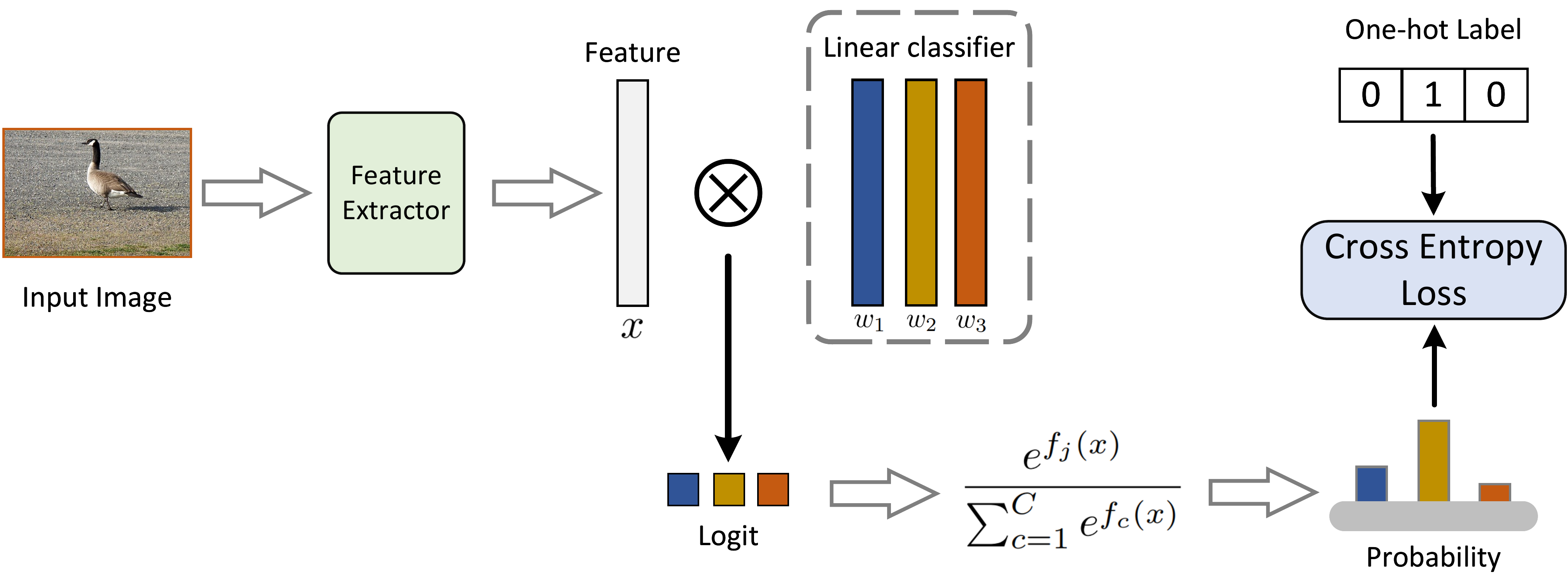} 
\caption{The overall training pipeline of vanilla linear classifier.} 
\label{vanilla-linear-classifier-flabel}
\end{figure*}

The general pipeline of the current image classification task is as follows: firstly, the features of the input samples are extracted using a deep neural network; secondly, the features are linearly transformed with a linear classifier to compute the similarity scores between the samples and the class center of each class; then, the similarity scores are modeled as a normalized posterior probability distribution using the softmax function; finally, the cross-entropy between the posterior probability and the class label is used as the loss function, as shown in Figure \ref{vanilla-linear-classifier-flabel}. This process can be rephrased as: a proxy class center (Figure \ref{MultiCenterImageEg-flabel}(a)) for each class is provided by employing a linear classifier and softmax cross-entropy is used to optimize the distance between features and their corresponding class centers. However, in real-world, one class can contain several local clusters (\emph{e.g.}, birds of different poses), making it challenging for a single class center to capture local structural diversity within that particular class. This limitation hampers existing methods' performance.

To capture the semantic relations between samples within a class, several researchers have proposed the utilization of multiple sub-centers to adapt to the complexity of data distribution \cite{qian2019softtriple,Zhao-2023-averageprecision}. These methods are commonly employed in deep metric learning (DML) and validated on fine-grained data (\emph{e.g.}, CUB200-2011 \cite{wah2011caltech}, Cars196 \cite{krause2015fine}), with unknown performance on large-scale image datasets (\emph{e.g.}, ImageNet \cite{5206848}). Moreover, these methods require more complex architectural designs and pose challenges when combined with various data augmentation strategies and softmax variants.

In this work, the proposed method aims to enhance the vanilla linear classifier by introducing multiple sub-centers, resulting in a novel classifier referred to as the multi-center classifier. In comparison to the vanilla linear classifier, the multi-center classifier exhibits improved capability in capturing the underlying data distribution. Specifically, we create a conditional Gaussian distribution for each class using the class center of the linear classifier as the mean and setting the variance as learnable parameters. Multiple sub-centers are then sampled from the conditional Gaussian distribution to extend the linear classifier with multiple centers for each class. To adapt to the multi-center classifier, we propose a label distribution method called Multi-Center Class Label, which ensures that each generated sub-center is involved in the training while the original class center dominates the label. The modification made by our multi-center classifier is limited to increasing the number of classes during the training phase. This allows for easy combination with current widely used data augmentations and various softmax variants without additional structural modifications. In addition, at test time the model removes the sampled sub-centers and only retains the mean of the conditional Gaussian distribution as the class center. This follows the vanilla linear classifier outputs, thus eliminating any need for extra parameters or computational overhead.

Extensive experiments show that multi-center classifier introduces higher feature diversity, reduces over-clustering, and learns more diverse class-distributions than vanilla linear classifier. In ImageNet, our multi-center classifier improves the top-1 accuracy of the original ResNet50 \cite{He-2016-CVPR} and Swin-T \cite{Liu-2021-ICCV} by +0.9\% and +0.4\%, respectively.

\begin{figure*}[t]
\centering
\includegraphics[width=0.9\linewidth]{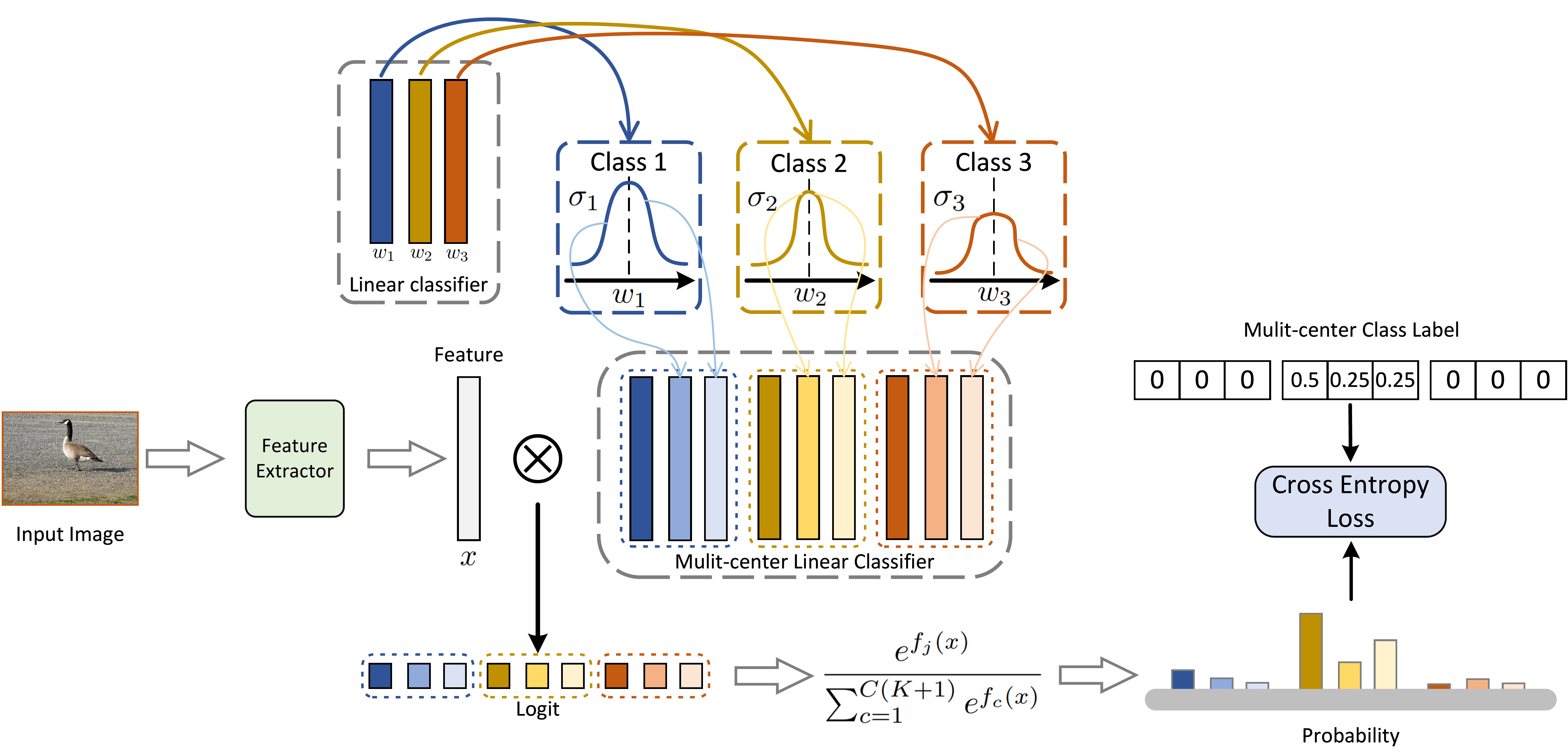} 
\caption{The overall training pipeline of the multi-center classifier. The weight ${{w}_{c}}$ of the linear classifier is used as the mean to create a conditional Gaussian distribution for each class. Multiple sub-centers are then sampled, and instead of using the original one-hot label, the cross-entropy loss is calculated based on the multi-center class label.}
\label{Multi-Center-classifier-flabel}
\end{figure*}

\section{Related Work}

Sub-center is usually introduced into the training process of image classification models to promote the model to learn more diverse class distributions. Softtriple \cite{qian2019softtriple} uses sub-centers in fine-grained image classification and these sub-centers help the neural network better capture the different modalities of the data. In sub-center arcface \cite{deng2020sub}, the sub-center is used to capture noise samples from face datasets for denoising to improve the performance of face datasets with much noise. NWDP \cite{10.1145/3474085.3475536} proposes to use multi-subcenters to distinguish in- and out-of-distribution noisy samples, and purify the web training data by discarding out-of-distribution noisy images and relabeling in-distribution images for better robustness and performance. MSCLDL \cite{9684999} proposed a Salient Object Detection method to decompose two classes into a certain number of sub-classes and the saliency value of each image region could be computed via exploring the relationship class and subclass labels. McSAP \cite{Zhao-2023-averageprecision} proposes multi-centers softmax reciprocal average precision loss to jointly supervise the learning of models by softmax with multi-centers and a ranking-based metric loss. McSAP allows to obtain a more optimally structured feature space with higher purity around the center.

Another technical route to improve feature diversity is to introduce Gaussian distribution into feature modeling. L-GM \cite{Wan-2018-CVPR} proposes a large-margin Gaussian Mixture loss that promotes both a high classification performance and an accurate modeling of the training feature distribution by involving a classification margin and a likelihood regularization. NIR \cite{Roth-2022-CVPR} proposes non-isotropy regularization for proxy-based deep metric learning. By leveraging Normalizing Flows, NIR enforces unique translatability of samples from their respective class proxies, allowing better learning of local structure for proxy-based objectives. Based on the concept of predefined evenly-distributed class centroids, POD \cite{9912430} proposes a loss function based on predefined optimal-distribution of latent features. The loss function restricts the norm-adaptive cosine distance between the latent feature vector of the sample and the predefined uniform class center. SALGL \cite{Zhu-2023-ICCV} proposes a novel scene-aware label graph learning framework, which is capable of learning visual representations for labels while fully perceiving their co-occurrence relationships under variable scenes towards accurate label prediction.

These methods focus on modeling the output features with Gaussian distributions, which requires complex design of the softmax function or even introducing a new loss function. However, our method incorporates the Gaussian distribution assumption into the classifier while still using the softmax loss function, thus avoiding complex loss function design. This property allows our method to be easily combined with various existing data augmentation strategies and softmax variants.

\section{Preliminaries}

Before introducing the proposed method, we briefly introduce the terminology and notation of vanilla linear classifier.

Considering a $C$ class classification task in which the softmax loss is used. The depth feature of the sample is denoted as $x$. The vanilla linear classifier $F$ can be viewed as a linear transformation layer (as shown in Figure \ref{vanilla-linear-classifier-flabel}), its learnable parameters are $\left[ {{w}_{1}},\cdots ,{{w}_{C}} \right]\in {{R}^{d\times C}}$, $d$ denotes the dimensionality of the feature. Calculate the posterior probability that $x$ belongs to $j$-th class by SoftMax as:
\begin{equation}
p\left( j|x \right)=\frac{{{e}^{{{f}_{j}}\left( x \right)}}}{\sum\nolimits_{c=1}^{C}{{{e}^{{{f}_{c}}\left( x \right)}}}}
\label{vanilla-softmax}
\end{equation}

\begin{equation}
{{f}_{c}}\left( x \right)=w_{c}^{T}x
\label{linear-weight}
\end{equation}
where $j\in \left[ 1,C \right]$, ${{f}_{c}}\left( x \right)$ is computed through inner product and the output represents the similarity (Logit) between the feature vector $x$ and the class center ${{w}_{c}}$. A larger value of the similarity ${{f}_{c}}\left( x \right)$ indicates a higher posterior probability of $x$ belonging to the class $c$.

We denote the features of the $i$-th sample as ${{x}_{i}}$ and the class label as ${{t}_{i}}$, and ${{t}_{i}}$ is then represented by the one-hot distribution ${{\tau }_{i}}$. Finally, compute the cross-entropy between the posterior probability and the class label $\tau $ as the loss function:
\begin{equation}
\mathcal{L}=-\sum\limits_{c=1}^{C}{{{\tau }_{c}}\log \left( {{p}_{c}} \right)}
\label{vanilla-cross-entropy}
\end{equation}

\begin{equation}
{{\tau }_{c}}=\left\{ \begin{matrix}
   0 & c\ne {{t}_{i}}  \\
   1 & c={{t}_{i}}  \\
\end{matrix} \right.
\label{vanilla-classLabel}
\end{equation}
where ${{p}_{c}}$ denotes the probability that the model predicts that ${{x}_{i}}$ belongs to the $c$-th class.

\section{Proposed Method}

In this section, we describe the proposed method in detail. Firstly, we elaborate on the process of sample sub-centers. Subsequently, we introduce the proposed multi-center class label. Lastly, we describe the different settings of the multi-center classifier during both training phase and testing phase.

\subsection{Sampling Sub-Center}

We create a conditional Gaussian distribution for each class using the class center $[{{w}_{1}},\cdots ,{{w}_{C}}]\in {{R}^{d\times C}}$ of the linear classifier as the mean ${{\mu }_{c}}\in {{R}^{d}}$ and setting the standard deviation ${{\sigma }_{c}}\in {{R}^{d}}$ as learnable parameters:
\begin{equation}
w_{c}^{(k)}\sim \mathcal{N}\left( {{w}_{c}},\sigma _{c}^{2}\mathbf{I}  \right)
\label{conditional-Gaussian-distribution}
\end{equation}

\begin{equation}
\begin{matrix}
   w_{c}^{(k)}={{w}_{c}}+{{\sigma }_{c}}\odot \varepsilon , & \varepsilon \sim \mathcal{N}\left( 0,\mathbf{I}  \right)  \\
\end{matrix}
\label{reparameterization}
\end{equation}
where $w_{c}^{(k)}$ denotes the $k$-th sub-center of class $c$, $k\in \left[ 1,K \right]$. $K$ denotes the number of sub-centers sampled, which is a hyperparameter (set to 2 by default), and $\odot $ is the element-wise product. Since the sampling process is not derivable, we sample $w_{c}^{(k)}$ using the reparameterization trick, as shown in Eq.(\ref{reparameterization}). We combine the original class centers with the sampled sub-centers to construct the multi-center classifier:
\begin{equation}
{{F}_{m}}=\left[ {{w}_{1}},w_{1}^{(1)}\cdots w_{1}^{(K)},\cdots ,{{w}_{C}},w_{C}^{(1)}\cdots w_{C}^{(K)} \right]
\label{multi-center-classifier-weight}
\end{equation}

\subsection{Multi-Center Class Label}

To achieve uniformity and simplicity in notational representation, we can treat the original $C$ class classification task as a classification task extended to $C\left( K+1 \right)$ after constructing the multi-center classifier. The learnable parameters of the multi-center classifier ${{F}_{m}}$ can be expressed as $[{{w}_{1}},\cdots ,{{w}_{C\left( K+1 \right)}}]\in {{R}^{d\times C\left( K+1 \right)}}$, as shown in Figure \ref{Multi-Center-classifier-flabel}. Thus computing $p\left( j|x \right)$ in our multi-center classifier is the same as the Eq.(\ref{vanilla-softmax}), except that $C$ is replaced by $C\left( K+1 \right)$. 

As the number of classes increases, the class label changes, ${{x}_{i}}$'s class label ${{t}_{i}}$ extends to ${{t}_{i}}\left( K+1 \right)$, and its sub-center label is $t_{i}^{(k)}={{t}_{i}}+k,k\in \left[ 1,K \right]$. For example, ${{x}_{i}}$'s class label ${{t}_{i}}=1$, for a $K=2$ multi-center classifier, ${{t}_{i}}=1\times \left( 2+1 \right)=3$, and the sub-center label $t_{i}^{(1)}=3+1=4$, $t_{i}^{(2)}=3+2=5$. In order to adapt to the increase in the number of classes due to the multi-center classifier, we propose a label distribution method called Multi-Center Class Label:
\begin{equation}
{{\tau }_{c}}=\left\{ \begin{matrix}
   \begin{matrix}
   0  \\
   0.5  \\
   \left( 1-0.5 \right)/K  \\
\end{matrix} & \begin{array}{*{35}{l}}
   c\ne {{t}_{i}},c\ne t_{i}^{(k)}  \\
   c={{t}_{i}}  \\
   c=t_{i}^{(k)}  \\
\end{array}  \\
\end{matrix} \right.
\label{multi-center-label}
\end{equation}

In multi-center class label, the original class center label is set to $0.5$ and the sub-center label is set to $\left( 1-0.5 \right)/K$. This ensures that every generated sub-center is involved in the training while the original class center dominates the label.

Similar to Eq.(\ref{vanilla-cross-entropy}), calculate the cross-entropy between the posterior probability and the mulit-center class label $\tau $ as the loss function:
\begin{equation}
{{\mathcal{L}}_{m}}=-\sum\limits_{c=1}^{C\left( K+1 \right)}{{{\tau }_{c}}\log \left( {{p}_{c}} \right)}
\label{multi-center-cross-entropy}
\end{equation}

We follow the VAE \cite{kingma2022autoencoding} and train the standard deviation ${{\sigma }_{c}}$ by calculating the KL-divergence between the generated Gaussian distribution and the standard normal distribution. Different from the loss function in the original VAE, the mean of the conditional Gaussian distribution in the multi-center classifier is equal to ${{w}_{c}}$, so only the standard deviation loss needs to be retained in the original loss:
\begin{equation}
{{\mathcal{L}}_{{{\sigma }^{2}}}}=-\frac{1}{2}\sum\limits_{j=1}^{d}{\left( 1+\log \sigma _{\left( j \right)}^{2}-\sigma _{\left( j \right)}^{2} \right)}
\label{KL-divergence-loss}
\end{equation}
where $d$ is the dimension of the variance vector and $\sigma _{\left( j \right)}^{2}$ represents the $j$-th component of the variance vector. Refer to VAE for the derivation of Eq.(\ref{KL-divergence-loss}).

The loss function used in our model is summarized as follows:
\begin{equation}
\mathcal{L}={{\mathcal{L}}_{m}}+{{\mathcal{L}}_{{{\sigma }^{2}}}}
\label{final-loss}
\end{equation}

The complete training process of the multi-center classifier is shown in Figure \ref{Multi-Center-classifier-flabel}.

\subsection{Testing Phase}

After completing the training process, the model discards the sampled sub-centers and retains only the mean of the conditional Gaussian distribution as the class center. Consequently, the multi-center classifier transforms into a conventional vanilla linear classifier. As a result, during testing, the model's structure remains identical to that of a vanilla classification model described in Section 3, without requiring any additional parameters or computational overhead.

\section{Experiments}

To demonstrate the effectiveness of the multi-center classifier, we combined it with the current mainstream convolutional neural network (CNN) and vision transformer (ViT) models for training. We conducted experiments on ImageNet-1K \cite{5206848}, as well as the small datasets Cifar-100 \cite{krizhevsky2009learning} and Mini-ImageNet \cite{krizhevsky2012imagenet}, for image classification. Based on the results of these experiments, we further evaluated the performance of the multi-center classifier when combined with various data augmentation methods and softmax variants. Finally, we conducted comprehensive ablation studies to analyze each component of the multi-center classifier.

\begin{table}[t]
   \centering
   \resizebox{\linewidth}{!}{
   \begin{tabular}{c|l|cc|c}
      \toprule 
       & Model   & Params & FLOPs  & Top-1 acc.           \\
      \midrule 
	   &ResNet-50 \cite{He-2016-CVPR}   & 23M   & 3.9G     &76.8    \\
       & RegNetY-4G \cite{Radosavovic-2020-CVPR}    & 21M   & 4.0G     &80.0    \\
       & DeiT-S \cite{pmlr-v139-touvron21a}   & 22M   & 4.6G     &79.8    \\
       & Swin-T \cite{Liu-2021-ICCV}    & 29M   & 4.5G     &81.3    \\
	  &SGFormer-S \cite{Ren-2023-ICCV}  & 22M   & 4.8G     &83.2     \\
      \midrule 
	 \multirow{5}{*}{MC}    
         &ResNet-50  & 25M   & 3.94G     &77.7 (\textbf{+0.9})    \\
       & RegNetY-4G  & 23M   & 4.04G     &80.8 (\textbf{+0.8})    \\
       & DeiT-S     & 23.5M   & 4.62G     &80.4 (\textbf{+0.6})    \\
       & Swin-T     & 30.5M   & 4.52G     &81.7  (\textbf{+0.4})   \\
	   &SGFormer-S   & 23.5M   & 4.82G     &83.5    (\textbf{+0.3}) \\
	\midrule 
	\midrule 
	 & ResNet-101 \cite{He-2016-CVPR}    & 45M   & 7.9G     &78.0    \\
	  &RegNetY-16G \cite{Radosavovic-2020-CVPR}    & 84M   & 16.0G     &82.9    \\
       &DeiT-B \cite{pmlr-v139-touvron21a}    & 86M   & 17.5G     &81.8    \\
       &Swin-B \cite{Liu-2021-ICCV}   & 88M   & 15.4G     &83.3     \\
 	 & SGFormer-B \cite{Ren-2023-ICCV}   & 78M   & 15.6G     &84.7     \\
	\midrule 
	\multirow{5}{*}{MC}  
	 & ResNet-101   & 47M   & 7.94G     &78.7  (\textbf{+0.7})  \\
	  &RegNetY-16G     & 86M   & 16.04G     &83.5  (\textbf{+0.6})  \\
       &DeiT-B    & 87.5M   & 17.52G     &82.2  (\textbf{+0.4})  \\
       &Swin-B   & 89.5M   & 15.42G     &83.6  (\textbf{+0.3})   \\
 	 & SGFormer-B    & 79.5M   & 15.26G     &85.1   (\textbf{+0.4})   \\
     \bottomrule 
   \end{tabular}
   }
  \caption{Comparison of different models on ImageNet-1K. \textbf{MC} denotes the use of multi-center classifier instead of the linear classifier used in the original model.}
   \label{ImageNet-Top1}
\end{table}

\begin{table}[t]
   \centering
   \resizebox{\linewidth}{!}{
   \begin{tabular}{c|l|cc|c}
       \toprule
       &Method   & Params & FLOPs                & Top-1 acc.           \\
       \midrule
	  &ResNet-50 \cite{He-2016-CVPR}   & 23M   & 3.9G     &78.2    \\
       & RegNetY-4G \cite{Radosavovic-2020-CVPR}    & 21M   & 4.0G     &78.3   \\
       & Swin-T \cite{Liu-2021-ICCV}    & 29M   & 4.5G     &78.8    \\
       & SGFormer-S \cite{Ren-2023-ICCV}   & 22M   & 4.8G     &82.3     \\
	\midrule
	\multirow{4}{*}{MC} 
	   &ResNet-50    & 25M   & 3.94G     &78.4  (\textbf{+0.2})   \\
       & RegNetY-4G     & 23M   & 4.04G     &78.6 (\textbf{+0.3})    \\
	 & Swin-T     & 30.5M   & 4.52G     &78.9 (\textbf{+0.1})   \\
       & SGFormer-S    & 23.5M   & 4.82G     &82.5 (\textbf{+0.2})    \\
       \midrule
       \midrule
	 & ResNet-101 \cite{He-2016-CVPR}    & 45M   & 7.9G     &78.7    \\
	  &RegNetY-16G \cite{Radosavovic-2020-CVPR}    & 84M   & 16.0G     &79.0    \\
       &Swin-B \cite{Liu-2021-ICCV}    & 88M   & 15.4G     &79.2     \\
 	 & SGFormer-B \cite{Ren-2023-ICCV}    & 78M   & 15.6G     &82.7     \\
	 \midrule
	\multirow{4}{*}{MC} 
	 & ResNet-101     & 47M   & 7.94G     &78.9 (\textbf{+0.2})   \\
	  &RegNetY-16G    & 86M   & 16.04G     &79.3  (\textbf{+0.3})  \\
       &Swin-B    & 89.5M   & 15.42G     &79.3   (\textbf{+0.1})  \\
 	 & SGFormer-B     & 79.5M   & 15.62G     &82.7   (\textbf{+0.0})  \\
       \bottomrule
   \end{tabular}
   }
   \caption{Comparison of different models on Cifar-100.}
   \label{Cifar-100-Top1}
\end{table}

\subsection{Classiﬁcation on the ImageNet-1K}

\noindent \textbf{Implementation details.} This setting mostly follows \cite{Liu-2021-ICCV}. We use the PyTorch toolbox \cite{paszke2019pytorch} to implement all our experiments. We employ an AdamW \cite{kingma2014adam} optimizer for 300 epochs using a cosine decay learning rate scheduler and 20 epochs of linear warm-up. A batch size of 1024, an initial learning rate of 0.001, and a weight decay of 0.05 are used. The image size is 224×224. We include most of the augmentation and regularization strategies of Swin transformer\cite{Liu-2021-ICCV} in training.

\noindent \textbf{Results.} Table \ref{ImageNet-Top1} compares the performance of CNN and ViT backbones on ImageNet-1K with and without the multi-center classifier. The experimental results show that in the CNN model, ResNet-50 and RegNetY-4G accuracy improved by +0.9\% and +0.8\% respectively after applying the multi-center classifier. Similarly, in the ViT model, the accuracy of Swin-T and SGFormer-S improved by +0.4\% and +0.3\%, respectively. The multi-center classifier is also effective for the base model, improving the accuracy of each model. In the ImageNet-1K experiment, we found that the multi-center classifier significantly improves the CNN model compared to the ViT model. This could be attributed to the fact that the CNN model (in the table) outputs 2048-dimensional features, providing a richer amount of information than the 768-dimensional output features of the ViT model, thus better capturing the intra-class local structure. To ensure a fair comparison, we did not adjust the ViT model for the multi-center classifier in our experiments.

\subsection{Classiﬁcation on Cifar-100 and Mini-ImageNet}

\noindent \textbf{Implementation details.} Follow the experimental settings in the above subsection.

\noindent \textbf{Results.} In Table \ref{Cifar-100-Top1} and Table \ref{Mini-ImageNet-Top1}, we compare the performance of the proposed multi-center classifier in combination with various models on small datasets. The experimental results show that although the multi-center classifier can effectively improve the performance of the model on small datasets, its improvement effect is not significant compared to large-scale datasets such as ImageNet. This indicates that large-scale data helps the multi-center classifier to learn more diverse class-distributions.

\begin{table}[t]
   \centering
   \resizebox{\linewidth}{!}{
   \begin{tabular}{c|l|cc|c}
       \toprule
     & Method   & Param. & FLOPs                & Top-1 acc.           \\
       \midrule
	  &ResNet-50 \cite{He-2016-CVPR}   & 23M   & 3.9G     &80.1    \\
       & RegNetY-4G \cite{Radosavovic-2020-CVPR}    & 21M   & 4.0G     &81.5   \\
      &Swin-T \cite{Liu-2021-ICCV}   & 29M   & 4.5G     &82.1    \\
      &SGFormer-S \cite{Ren-2023-ICCV}    & 22M   & 4.8G     &84.4     \\
	 \midrule
	\multirow{4}{*}{MC} 
	  &ResNet-50   & 25M   & 3.94G     &80.5    (\textbf{+0.4}) \\
       & RegNetY-4G     & 23M   & 4.04G     &81.8   (\textbf{+0.3}) \\
      &Swin-T  & 30.5M   & 4.52G     &82.4    (\textbf{+0.3}) \\
      &SGFormer-S     & 23.5M   & 4.82G     &84.9     (\textbf{+0.5}) \\
      \midrule
	\midrule
	 & ResNet-101 \cite{He-2016-CVPR}    & 45M   & 7.9G     &81.0    \\
	  &RegNetY-16G \cite{Radosavovic-2020-CVPR}    & 84M   & 16.0G     &82.1    \\
      &Swin-B \cite{Liu-2021-ICCV}    & 88M   & 15.4G     &82.3     \\
 	& SGFormer-B \cite{Ren-2023-ICCV}   & 78M   & 15.6G     &84.6     \\
	\midrule
	\multirow{4}{*}{MC} 
	 & ResNet-101     & 47M   & 7.94G     &81.3  (\textbf{+0.3})  \\
	  &RegNetY-16G    & 86M   & 16.04G     &82.5  (\textbf{+0.4})  \\
      &Swin-B    & 89.5M   & 15.42G     &82.5   (\textbf{+0.2})  \\
 	& SGFormer-B    & 79.5M   & 15.62G     &84.9   (\textbf{+0.3})  \\
      \bottomrule
   \end{tabular}
   }
   \caption{Comparison of different models on Mini-ImageNet.}
   \label{Mini-ImageNet-Top1}
\end{table}

\subsection{Combining Data Augmentations and Softmax Variants}

The modification made by our multi-center classifier is limited to increasing the number of classes during the training phase, so it can be easily combined with current widely used data augmentations and various softmax variants.

\begin{table}[t]
   \centering
   \resizebox{\linewidth}{!}{
   \begin{tabular}{c|l|cc|c}
      \toprule 
       & Model   & Params & FLOPs  & Top-1 acc.           \\
      \midrule 
	  &ResNet-50 \cite{He-2016-CVPR}   & 23M   & 3.9G     &76.5    \\
       & Swin-T \cite{Liu-2021-ICCV}    & 29M   & 4.5G     &79.3    \\
 	\midrule 
	 \multirow{2}{*}{LS}    
       &ResNet-50  & 23M   & 3.9G     &76.8     \\
       & Swin-T     & 29M   & 4.5G     &79.9     \\
 	\midrule 
	 \multirow{2}{*}{MC}    
       &ResNet-50  & 25M   & 3.94G     &77.7     \\
       & Swin-T     & 30.5M   & 4.52G     &80.5     \\
      \midrule 
	 \multirow{2}{*}{MC + LS}    
       &ResNet-50  & 25M   & 3.94G      &\textbf{77.7}     \\
       & Swin-T     & 30.5M   & 4.52G     &\textbf{80.6}     \\
     \bottomrule 
   \end{tabular}
   }
  \caption{Performance of multi-center classifier combined with Label Smoothing on ImageNet-1K. \textbf{LS} denotes Label Smoothing. \textbf{MC} denotes Multi-Center classifier.}
   \label{Label-Smoothing}
\end{table}

\begin{table}[t]
   \centering
   \resizebox{\linewidth}{!}{
   \begin{tabular}{c|l|cc|c}
      \toprule 
       & Model   & Params & FLOPs  & Top-1 acc.           \\
      \midrule 
	  &ResNet-50 \cite{He-2016-CVPR}   & 23M   & 3.9G     &76.5    \\
       & Swin-T \cite{Liu-2021-ICCV}    & 29M   & 4.5G     &79.3    \\
 	\midrule 
	 \multirow{2}{*}{MixUp}    
       &ResNet-50  & 23M   & 3.9G     &77.2     \\
       & Swin-T     & 29M   & 4.5G     &81.3     \\
 	\midrule 
	 \multirow{2}{*}{MC}    
       &ResNet-50  & 25M   & 3.94G     &\textbf{77.7}     \\
       & Swin-T     & 30.5M   & 4.52G     &80.5     \\
      \midrule 
	 \multirow{2}{*}{MC + MixUp}    
       &ResNet-50  & 25M   & 3.94G      &77.6     \\
       & Swin-T     & 30.5M   & 4.52G     &\textbf{81.7}     \\
     \bottomrule 
   \end{tabular}
   }
  \caption{Performance of multi-center classifier combined with MixUp on ImageNet-1K.}
   \label{MixUp}
\end{table}

\begin{table}[h]
   \centering
   \resizebox{\linewidth}{!}{
   \begin{tabular}{c|l|cc|c}
      \toprule 
       & Model   & Params & FLOPs  & Top-1 acc.           \\
      \midrule 
	 \multirow{2}{*}{MC + Softmax} 
	  &ResNet-50 \cite{He-2016-CVPR}   & 25M   & 3.94G     &77.7    \\
       & Swin-T \cite{Liu-2021-ICCV}    & 30.5M   & 4.52G     &\textbf{81.7}    \\
 	\midrule 
	 \multirow{2}{*}{MC + L-Softmax \cite{liu2017largemargin}}    
       &ResNet-50  & 25M   & 3.94G     &77.3     \\
       & Swin-T     & 30.5M   & 4.52G     &81.1     \\
      \midrule 
	 \multirow{2}{*}{MC + L-AM \cite{Deng-2019-CVPR}}    
       &ResNet-50  & 25M   & 3.94G      &\textbf{77.8}     \\
       & Swin-T     & 30.5M   & 4.52G     &81.6     \\
     \bottomrule 
   \end{tabular}
   }
  \caption{Performance of multi-center classifier combined with softmax variants on ImageNet-1K.}
   \label{softmax-variants}
\end{table}

\subsubsection{Data augmentations}

Label Smoothing \cite{Szegedy-2016-CVPR} adds noise to labels to prevent the model from predicting labels too confidently during training, which improves the generalization ability of the model. MixUp alternates between cutmix \cite{Yun-2019-ICCV} and cutout \cite{zhang2018mixup}. It is the default data augmentation for swin ViT \cite{Liu-2021-ICCV}. Tables \ref{Label-Smoothing} and \ref{MixUp} show the performance of the two data augmentation methods used alone and in combination with the multi-center classifier. The experimental results indicate that our multi-center classifier can be combined with commonly used data augmentation methods to further improve the model's performance.

\subsubsection{Softmax variants}

In Table \ref{softmax-variants}, we show the performance of the two softmax variants when combined with the multi-center classifier. The experimental results demonstrate that our multi-center classifier can be trained in combination with these softmax variants without complex structural modifications.

\begin{figure}[h]
\centering
\includegraphics[width=0.9\linewidth]{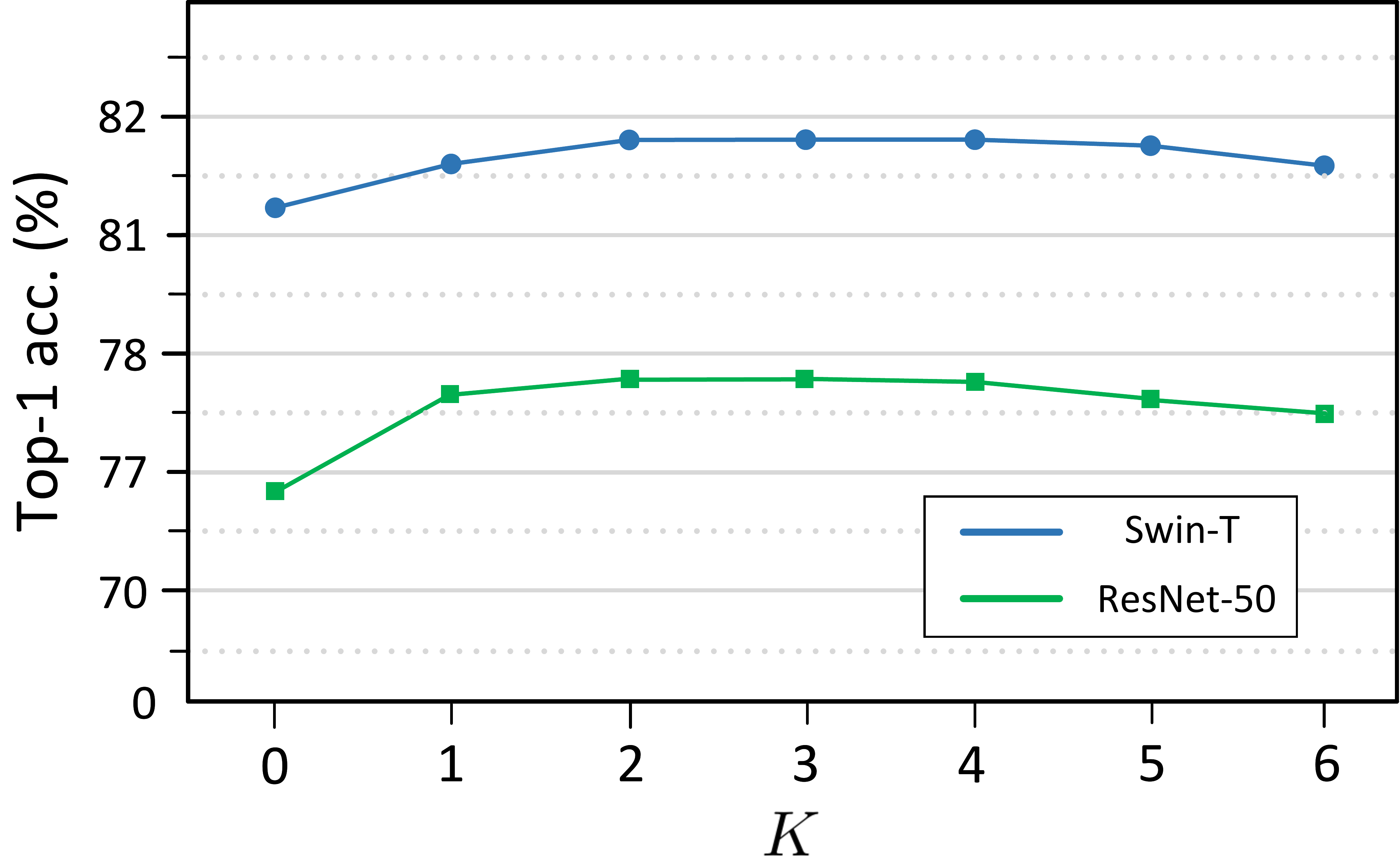} 
\caption{Effect of the number of sub-centers on model performance on ImageNet-1K.}
\label{K-Number-flabel}
\end{figure}

\subsection{Ablation Study}

We perform ablation studies on image classification for the fundamental designs of our multi-center classifier. For a fair comparison, we only change one component for each ablation.

\subsubsection{Number of sub-centers}

In this subsection, we investigate the impact of the number of sub-center $K$ on the performance of the multi-center classifier. As shown in Figure \ref{K-Number-flabel}, sampling a small number of sub-centers is sufficient to support the effective operation of the multi-center classifier. This is attributed to the fact that sub-centers are re-sampled with each forward propagation, so we don't need to sample a large number of sub-centers each time to model the feature distribution, as long as the training epoch is enough.

\begin{table}[t]
   \centering
   \resizebox{\linewidth}{!}{
   \begin{tabular}{c|l|cc|c}
      \toprule 
       & Model   & Params & FLOPs  & Top-1 acc.           \\
      \midrule 
	 \multirow{2}{*}{MC (without ${{\mathcal{L}}_{{{\sigma }^{2}}}}$)} 
	  &ResNet-50 \cite{He-2016-CVPR}   & 25M   & 3.94G     &77.0    \\
       & Swin-T \cite{Liu-2021-ICCV}    & 30.5M   & 4.52G     &81.4    \\
 	\midrule 
	 \multirow{2}{*}{MC + ${{\mathcal{L}}_{{{\sigma }^{2}}}}$ }    
       &ResNet-50  & 25M   & 3.94G      &\textbf{77.7}     \\
       & Swin-T     & 30.5M   & 4.52G     &\textbf{81.7}     \\
     \bottomrule 
   \end{tabular}
   }
  \caption{Effect of the standard deviation loss ${{\mathcal{L}}_{{{\sigma }^{2}}}}$ on model performance on ImageNet-1K.}
   \label{standard-deviation-loss}
\end{table}

\subsubsection{Standard deviation loss}

In this subsection, we explore the effect of standard deviation loss ${{\mathcal{L}}_{{{\sigma }^{2}}}}$ on the performance of the multi-center classifier. As shown in Table \ref{standard-deviation-loss}, when the standard deviation loss is removed from the model resulted in a significant decrease in performance. This result proves the importance of standard deviation loss in our multi-center classifier. The loss function forces the model to learn a larger standard deviation of the Gaussian distribution, which reduces over-clustering and facilitates the model to learn more diverse distribution of intra-class features.

\section{Conclusion}

To mitigate the issue of over-clustering and enhance the diversity of feature distributions, we propose a multi-center classifier based on the assumption that deep features in the training data follow a Gaussian Mixture distribution. For each class, we create a conditional Gaussian distribution to generate multiple centers. To accommodate this multi-center classifier, we propose a label distribution method called Multi-Center Class Label, which ensures that every generated sub-center is involved in the training. Our approach can be seamlessly integrates with various data augmentations and softmax variants, while requiring no additional parameters or computational overhead during testing. Extensive experiments on image classification show that the proposed multi-center classifier serves as an effective alternative to widely used linear classifiers.


\bibliographystyle{named}
\bibliography{ijcai24}

\end{document}